\documentclass{egpubl}
\usepackage{eg2024}
 
\ConferencePaper        
\CGFStandardLicense

\usepackage[T1]{fontenc}
\usepackage{dfadobe}  

\BibtexOrBiblatex
\electronicVersion
\PrintedOrElectronic

\ifpdf \usepackage[pdftex]{graphicx} \pdfcompresslevel=9
\else \usepackage[dvips]{graphicx} \fi

\usepackage{egweblnk} 
\usepackage{lineno}

\usepackage{amsmath}
\usepackage{amssymb}
\usepackage{bm}
\usepackage{xcolor}
\usepackage[ruled,linesnumbered]{algorithm2e}
\usepackage{booktabs}

\newcommand{\colorRef}[1]{\textcolor{black}{#1}} 
\usepackage[capitalize]{cleveref}
\crefname{figure}{\colorRef{Fig.}}{\colorRef{Figs.}}
\Crefname{figure}{\colorRef{Figure}}{\colorRef{Figures}}
\crefname{section}{\colorRef{Sec.}}{\colorRef{Secs.}}
\Crefname{section}{\colorRef{Section}}{\colorRef{Sections}}
\creflabelformat{equation}{#2\textup{#1}#3}
\crefname{equation}{\colorRef{Eq.}}{\colorRef{Eqs.}}
\Crefname{equation}{\colorRef{Equation}}{\colorRef{Equations}}
\Crefname{table}{\colorRef{Table}}{\colorRef{Tables}}
\crefname{table}{\colorRef{Tab.}}{\colorRef{Tabs.}}
\Crefname{table}{\colorRef{Table}}{\colorRef{Tables}}
\crefname{table}{\colorRef{Tab.}}{\colorRef{Tabs.}}
\crefname{algorithm}{\colorRef{Alg.}}{\colorRef{Algs}}
\Crefname{algorithm}{\colorRef{Algorithm}}{\colorRef{Algorithms}}

\LinesNumbered

\SetCommentSty{mycommfont}

\usepackage{caption}
\usepackage{subcaption}

\title[Learning to Stabilize Faces]{Learning to Stabilize Faces}

\author[J. Bednarik \& E. Wood \& V. Choutas \& T. Bolkart \& D. Wang \& C. Wu \& T. Beeler]
{\parbox{\textwidth}{\centering J. Bednarik and E. Wood and V. Choutas and T. Bolkart and D. Wang and C. Wu and T. Beeler
        }
        \\
{\parbox{\textwidth}{\centering Google
       }
}
}

\begin{document}

\newcommand{\hideablecomment}[1]{#1}
\newcommand{\hidecomments}{%
    \renewcommand{\hideablecomment}[1]{}%
}%

\renewcommand{\sectionautorefname}{Section}
\renewcommand{\subsectionautorefname}{Section}
\renewcommand{\subsubsectionautorefname}{Section}
\renewcommand{\figureautorefname}{Figure}

\definecolor{lime}{RGB}{175,175,0}
\definecolor{orange}{rgb}{1,0.5,0}

\newcommand{\jb}[1]{\hideablecomment{\textcolor{red}{\textbf{JB:} #1}}}
\newcommand{\thbe}[1]{\hideablecomment{\textcolor{green}{\textbf{ThBe:} #1}}}
\newcommand{\cw}[1]{\hideablecomment{\textcolor{blue}{\textbf{CW:} #1}}}
\newcommand{\ew}[1]{\hideablecomment{\textcolor{cyan}{\textbf{EW:} #1}}}
\newcommand{\tibo}[1]{\hideablecomment{\textcolor{magenta}{\textbf{TiBo:} #1}}}
\newcommand{\vc}[1]{\hideablecomment{\textcolor{lime}{\textbf{VC:} #1}}}
\newcommand{\dw}[1]{\hideablecomment{\textcolor{orange}{\textbf{DW:} #1}}}
\newcommand{\needscitation}{\hideablecomment{\textcolor{red}{[CITE]}}}

\newcommand{\minorrev}[1]{\textcolor{orange}{#1}}

\newcommand{\head}{H}
\newcommand{\headi}[1]{\head_{#1}}

\newcommand{\skin}{\text{skin}}
\newcommand{\skull}{\text{skull}}

\newcommand{\vertsskin}{V}
\newcommand{\vertsskull}{W}
\newcommand{\nvertsskin}{N_{\vertsskin}}
\newcommand{\nvertsskull}{N_{\vertsskull}}

\newcommand{\vertsskini}[1]{\vertsskin_{#1}}
\newcommand{\vertsskulli}[1]{\vertsskull_{#1}}

\newcommand{\matstab}{S}
\newcommand{\matstabstar}{S^{*}}
\newcommand{\funcstab}{\mathcal{S}}
\newcommand{\funcstabcall}[2]{\funcstab(\vertsskini{#1}, \vertsskini{#2})}
\newcommand{\matrigidtf}{Z}
\newcommand{\matrigidtfi}[1]{Z_{#1}}

\newcommand{\energystab}{\mathcal{E}}
\newcommand{\energystabcall}[3]{\energystab(\vertsskulli{#1}, \vertsskulli{#2}, #3)}

\newcommand{\threedmm}{\mathcal{M}}
\newcommand{\threedmmparams}{\Theta}
\newcommand{\threedmmdata}{\Psi}
\newcommand{\threedmmwithdata}{\threedmm_{\threedmmdata}}
\newcommand{\paramsid}{\beta}
\newcommand{\paramsexpr}{\phi}
\newcommand{\paramsrot}{\theta}
\newcommand{\paramstransl}{\tau}
\newcommand{\njoints}{K}
\newcommand{\nparamsid}{\lone{\paramsid}}
\newcommand{\nparamsexpr}{\lone{\paramsexpr}}
\newcommand{\nparamsrot}{\njoints \times 3}
\newcommand{\nparamstransl}{3}
\newcommand{\threedmmcallexpri}[1]{\threedmm(\paramsid, \paramsexpr_{#1}, \paramsrot, \paramstransl)}
\newcommand{\basesid}{\mathbf{I}}
\newcommand{\basesexpr}{\mathbf{E}}
\newcommand{\nbasesid}{\nparamsid}
\newcommand{\nbasesexpr}{\nparamsexpr}
\newcommand{\joint}{\mathbf{j}}
\newcommand{\jointneck}{\joint_{\text{neck}}}
\newcommand{\linearblendskinning}{\mathcal{L}}
\newcommand{\vertsskinbind}{\vertsskin_{\text{bind}}}
\newcommand{\lbsweights}{\mathbf{W}}
\newcommand{\threedmmtempl}{T}
\newcommand{\threedmmtemplfront}{\widehat{\threedmmtempl}}
\newcommand{\threedmmtempljoints}{J}
\newcommand{\basesidjoints}{\mathbf{Q}}
\newcommand{\jointparents}{P}
\newcommand{\skinningtransforms}{\mathbf{X}}
\newcommand{\jointsbind}{J_{\text{bind}}}

\newcommand{\jointbind}{\joint_{\text{bind}}}
\newcommand{\jointtempl}{\overline{\joint}}
\newcommand{\jointbasesid}{\mathbf{Q}}
\newcommand{\jointwithname}[1]{\joint_{\text{#1}}}
\newcommand{\jointtemplwithname}[1]{\jointtempl_{\text{#1}}}

\newcommand{\neckvector}{n}

\newcommand{\vertsneutral}{\vertsskin^{\text{neutral}}}
\newcommand{\neutralsubset}{\mathtt{N}}

\newcommand{\nsubjects}{2\,519}
\newcommand{\nexpressions}{57}
\newcommand{\nframes}{38\,360}
\newcommand{\nthreedmmskinverts}{17\,821}

\newcommand{\rotation}{R}
\newcommand{\translation}{\mathbf{t}}
\newcommand{\nnparams}{\Omega}
\newcommand{\nnfuncstab}{\funcstab_{\nnparams}}
\newcommand{\featureextractor}{\mathcal{F}}
\newcommand{\tfregressor}{\mathcal{R}}
\newcommand{\latentdim}{L}
\newcommand{\latent}{\mathbf{z}}
\newcommand{\latenti}[1]{\latent_{#1}}
\newcommand{\loss}{\mathcal{L}}
\newcommand{\lossrot}{\loss_{R}}
\newcommand{\losstransl}{\loss_{T}}
\newcommand{\weightlosstransl}{\alpha_{T}}
\newcommand{\gtrot}{\overline{\rotation}}
\newcommand{\gttransl}{\overline{\translation}}

\newcommand{\vertsfront}{\widetilde{V}}
\newcommand{\vertsfronti}[1]{\vertsfront_{#1}}

\newcommand{\vertsproctempl}{\widehat{V}}
\newcommand{\vertsproctempli}[1]{\vertsproctempl_{#1}}

\newcommand{\front}{C}
\newcommand{\nvertsfront}{N_{\front}}

\newcommand{\funcmaskfront}{\mathcal{C}}
\newcommand{\procrustes}{\mathcal{P}}

\newcommand{\datasetid}{\mathbf{B}}
\newcommand{\datasetexpr}{\bm{\Phi}}
\newcommand{\procrustesmat}{\mathcal{Q}}
\newcommand{\matfourbyfour}{Q \in \real^{4\times4}}
\newcommand{\funcsamplerigidtf}{\mathcal{G}}
\newcommand{\noiserot}{\epsilon_{R}}
\newcommand{\noisetransl}{\epsilon_{T}}
\newcommand{\noiseexpr}{\epsilon_{\paramsexpr}}
\newcommand{\gttf}{\overline{\matstab}}
\newcommand{\meanid}{\paramsid_{\mu}}
\newcommand{\stdid}{\paramsid_{\sigma}}
\newcommand{\tffromto}[2]{\matstab_{#1 \rightarrow #2}}
\newcommand{\tftargettemplate}{\tffromto{t}{\threedmmtemplfront}}
\newcommand{\tfsourcetarget}{\tffromto{s}{t}}
\newcommand{\randomrigidtf}{\matstab_{\epsilon}}
\newcommand{\randomrigidtfi}[1]{{\randomrigidtf}_{#1}}

\newcommand{\ngtsubjects}{15}
\newcommand{\ngtframes}{45}
\newcommand{\ngtstabilizationframes}{30}
\newcommand{\nvalidationsubjects}{5}
\newcommand{\ntestsubjects}{10}
\newcommand{\nvalidationframes}{10}
\newcommand{\ntestframes}{20}

\newcommand{\metric}{m}
\newcommand{\metricmeanvertdist}{\metric_{d}}
\newcommand{\metricmaxvertdist}{\metric_{x}}
\newcommand{\metricaucparag}{\metric_{AUC}}
\newcommand{\metricauc}{\metric_{\text{AUC}}}
\newcommand{\predtf}{\matstab}
\newcommand{\gtverts}{\overline{\vertsskin}}
\newcommand{\ntestsamples}{N}
\newcommand{\ntestverts}{M}
\newcommand{\ithsamplejthvert}[2]{\vertsskin^{(#1)}_{#2}}
\newcommand{\ithsamplejthvertpred}[2]{\ithsamplejthvert{#1}{#2}}
\newcommand{\ithsamplejthvertgt}[2]{\gtverts^{(#1)}_{#2}}
\newcommand{\pckmin}{0}
\newcommand{\pckmax}{5}

\newcommand{\our}{OUR}
\newcommand{\zerorotparams}{\mathbf{0}_{\paramsrot}}
\newcommand{\zerotranslparams}{\mathbf{0}_{\paramstransl}}
\newcommand{\methodprocrustes}{\text{PROC}}
\newcommand{\procrusteshead}{\methodprocrustes_{\text{head}}}
\newcommand{\procrustesface}{\methodprocrustes_{\text{face}}}
\newcommand{\procrustesupper}{\methodprocrustes_{\text{upper}}}
\newcommand{\unpose}{\text{UNPOSE}}
\newcommand{\unposesharedid}{\unpose_{\text{id}}}
\newcommand{\methodconfidencemap}{CMAP}

\newcommand{\headregion}{Head}
\newcommand{\faceregion}{Face}
\newcommand{\upperfaceregion}{Upper}

\newcommand{\fullmask}{Full}
\newcommand{\theatermask}{Face\&Neck\xspace}
\newcommand{\superheromask}{Superhero}

\newcommand{\real}{\mathbb{R}}
\newcommand{\integers}{\mathbb{Z}}
\newcommand{\transpose}[1]{#1^{\top}}
\newcommand{\lone}[1]{\lvert #1 \rvert}
\newcommand{\ltwo}[1]{\lVert #1 \rVert}
\newcommand{\frobenius}[1]{\ltwo{#1}_{\text{F}}}
\newcommand{\argmin}{\mathop{\mathrm{arg\,min}}}
\newcommand{\funccallone}[2]{#1\left(#2\right)}
\newcommand{\funccalltwo}[3]{#1\left(#2, #3\right)}
\newcommand{\funccalthree}[4]{#1\left(#2, #3, #4\right)}
\newcommand{\normaldistrib}[2]{\mathcal{N}(#1, #2)}
\newcommand{\uniformdistrib}{\mathcal{U}}
\newcommand{\uniformdistribone}[1]{\uniformdistrib(#1)}
\newcommand{\uniformdistribtwo}[2]{\uniformdistrib(#1, #2)}
\newcommand{\identitymat}{\mathbf{I}}
\newcommand{\identitymati}[1]{\identitymat_{#1}}
\newcommand{\zeros}{\mathbf{0}}
\newcommand{\ones}{\mathbf{1}}
\newcommand{\invert}[1]{#1^{-1}}
\newcommand{\diagonal}[1]{\text{Diag}(#1)}

\newcommand\thefont{\expandafter\string\the\font}

\newcommand{\rigidtf}{S_{\epsilon}}
\newcommand{\zerovector}{\mathbf{0}}
\newcommand{\randomangle}{\alpha}
\newcommand{\randomaxis}{\mathbf{a}}
\newcommand{\funcangleaxis}{\mathcal{A}}

\newcommand{\verts}{U}
\newcommand{\sourceverts}{\verts_{s}}
\newcommand{\targetverts}{\verts_{t}}
\newcommand{\maskedsourceverts}{\tilde{\verts}_{s}}
\newcommand{\maskedtargetverts}{\tilde{\verts}_{t}}
\newcommand{\weights}{\mathbf{w}}
\newcommand{\tiledweights}{W}
\newcommand{\hadamard}{\odot}
\newcommand{\nverts}{N}
\newcommand{\funcstd}{\sigma}
\newcommand{\neighborhoodindex}{k}
\newcommand{\neighborhood}{\mathbb{N}}
\newcommand{\onesvectorfour}{[1, 1, 1, 1]}

\newcommand{\losstermweight}{\alpha}
\newcommand{\alphadata}{\losstermweight_{\text{data}}}
\newcommand{\alphareg}{\losstermweight_{\text{reg}}}
\newcommand{\alphacontrast}{\losstermweight_{\sigma}}
\newcommand{\alphaconsist}{\losstermweight_{\neighborhood}}

\newcommand{\lossdata}{\loss_{\text{data}}}
\newcommand{\lossreg}{\loss_{\text{reg}}}
\newcommand{\losscontrast}{\loss_{\sigma}}
\newcommand{\lossconsist}{\loss_{\neighborhood}}

\newcommand{\cmaporiginal}{\emph{Original}}
\newcommand{\cmapcontrast}{\emph{Contrast}}
\newcommand{\cmapconstrastandconsistent}{\emph{Contrast\&Consistent}}

\teaser{
\includegraphics[width=\linewidth]{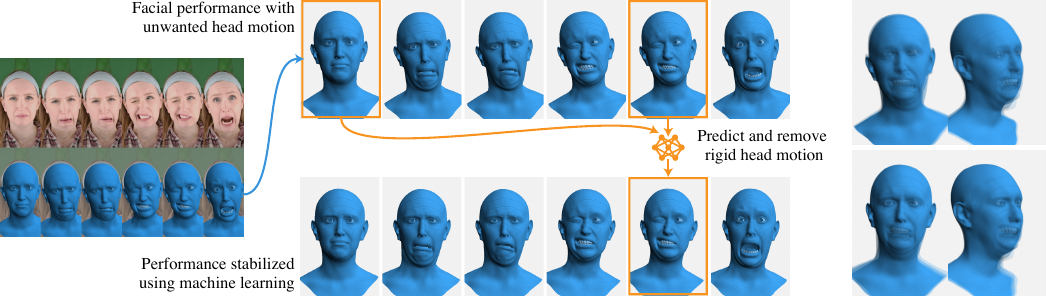}
\centering
\caption{\label{fig:teaser}
Our method stabilizes face meshes using machine learning.
Given an input pair of face meshes, we predict the rigid transform between the two, and remove it.
Shown on the right are all face meshes overlaid. After stabilization, the rigid head parts appear aligned.
}
}

\maketitle
\begin{abstract}
Nowadays, it is possible to scan faces and automatically register them with high quality.
However, the resulting face meshes often need further processing: we need to \emph{stabilize} them to remove unwanted head movement.
Stabilization is important for tasks like game development or movie making which require facial expressions to be cleanly separated from rigid head motion.
Since manual stabilization is labor-intensive, there have been attempts to automate it.
However, previous methods remain impractical: they either still require some manual input, produce imprecise alignments, rely on dubious heuristics and slow optimization, or assume a temporally ordered input.
Instead, we present a new learning-based approach that is simple
and fully automatic.
We treat stabilization as a regression problem: given two face meshes, our network directly predicts the rigid transform between them that brings their skulls into alignment.
We generate synthetic training data using a 3D Morphable Model (3DMM), exploiting the fact that 3DMM parameters separate skull motion from facial skin motion.
Through extensive experiments we show that our approach outperforms the state-of-the-art both quantitatively and qualitatively on the tasks of stabilizing discrete sets of facial expressions as well as dynamic facial performances.
Furthermore, we provide an ablation study detailing the design choices and best practices to help others adopt our approach for their own uses.
Supplementary videos can be found on the project webpage \url{syntec-research.github.io/FaceStab}.
\end{abstract}  
\section{Introduction} \label{sec:introduction}
High-fidelity face data is generally captured using multiple cameras, where
a 3D mesh of fixed topology is non-rigidly registered to a collection of images.
This is either done using traditional multi-view stereo and model-based optimization \cite{egger20}, or more modern deep learning techniques \cite{bolkart23,li2021,liu2022}.

Often, a capture session will involve a single subject performing multiple expressions.
When comparing meshes corresponding to different expressions, one observes inter-sample vertex motion caused by two phenomena.
First, the skin deforms as the subject contracts underlying facial muscles to perform expressions, and
second, the whole head rigidly translates and rotates as the subject moves their neck and body.
This rigid motion can be small, e.g. accidental shifts of the head in a sequence of static expressions, or large, e.g. dramatic head shakes in an expressive performance.
For many use cases, we are only interested in the former, with the latter representing undesirable residual motion that pollutes the data.

The goal of \emph{stabilization} is hence to remove unwanted rigid head motion, such as the one shown in \cref{fig:teaser}.
This is an essential step for
building personalized face rigs for animation \cite{alexander2009,seymour2017},
facial deformation transfer and retargeting \cite{bouaziz13,li13,chandran22}, and
creating data-driven linear bases for 3D Morphable Models (3DMM) \cite{egger20}.

Stabilization can be done manually by a skilled artist, but this is highly labor-intensive, so attempts have been made to automate it.
Automatic stabilization would be easy if some parts of the face never moved during expressions, but unfortunately, this is not the case.
The ultimate solution might track and align the skull itself, e.g. using X-Ray,
but since this is impractical we must do our best with the visible parts of the face.
Alas, high-quality fully automatic solutions are not yet readily available.
Existing solutions either rely on imprecise Procrustes alignment of the skin vertices \cite{weise11,bouaziz13,wu18,cao18},
require manually annotated landmarks \cite{beeler14,wu16},
have prohibitive constraints on the input format \cite{fyffe17,lamarre18},
or only produce approximate results by unposing a 3DMM \cite{li17}.

We present the first learning-based method for stabilization.
We treat stabilization as a regression problem: given an input pair of meshes, we use a neural network to predict the unwanted rigid transform between the two underlying skulls.
We train our network using a large and diverse synthetic dataset, generated by randomly sampling both a 3DMM and a database of registered faces.
Crucially, we exploit the fact that 3DMMs have a stable skull by design, which lets us freely recombine identity and expression into novel training samples.
We show that such a model trained on synthetic data works well on real face meshes and achieves state-of-the-art accuracy both visually and quantitatively.

\section{Related Work} \label{sec:related_work}
Being an essential step of many face synthesis and analysis tasks, stabilization has been studied both on its own and as a part of a larger problem.
Traditional approaches rely on rigid Procrustes alignment \cite{gower75} either directly, or combined with ICP \cite{arun1987least} when mesh correspondences are not known \cite{weise11,bouaziz13}
To prevent bad alignments often caused by jaw movement, these approaches typically only use the upper face region for alignment.

A different body of work achieves higher accuracy than Procrustes by considering the real physiology of the human skull and skin \cite{zoss19}.
\cite{beeler14, wu16} first estimate the shape of the underlying skull and optimize its pose via anatomically motivated heuristics involving skin thickness and skin-bone sliding.
While accurate, the methods are not fully automatic and require per-subject one-time skull shape initialization, which involves costly manual landmark annotation.

A fully automated method was proposed in \cite{lamarre18}, which assumes that given a reference coordinate frame of the skull, every facial surface point moves around its rest position. The authors rely on a heuristic linking the sharpness of the per-point position histogram to the stabilization quality and design a corresponding optimization scheme. The method requires an input in the form of a smooth facial performance and thus cannot be used to stabilize arbitrary expression pairs.

Head stabilization is often addressed as a step within face reconstruction and tracking pipelines \cite{fyffe17,wu18,cao18}. The method of \cite{fyffe17} initializes the stabilization using Procrustes alignment, followed by finding axis-aligned rotations of the reconstructed mesh that best explain the observation. This method relies on an optimization scheme and requires the availability of reconstructed eyeballs. \cite{wu18} convert the meshes to geometry images \cite{gu02} and show that Procrustes alignment can perform well as long as it is applied only on a proper region of the head which is learned in a data-driven fashion. Similarly, \cite{cao18} divide the template face mesh into segments contributing to the rigid stabilization optimization with different weights, which are a function of the input. Similarly to \cite{lamarre18}, the approach only works for facial performances and thus cannot be used to stabilize expression sets.

Other approaches use 3DMMs to stabilize faces. A common practice is to fit the model to the observed data and then use the model parameters to undo any undesirable global motion due to rigid transformation or neck rotation \cite{egger20,li17}. The main drawback is the sensitivity to the quality of the 3DMM fits. Any residual fitting error is arbitrarily distributed between the model parameters and the global rigid motion. Furthermore, 3DMM parameters underlying the registered meshes are not always available, which is typically the case for modern deep-learning-based solutions \cite{bolkart23,li2021,liu2022}.

In contrast to prior work, our approach is fully automatic.
We do not require a temporally consistent sequence of frames; our method works on just two meshes at a time.
Finally, we only use a 3DMM to train our method; at inference time we require only 3D meshes.

\section{Methodology} \label{sec:methodology}

In \cref{ssec:problem_fomulation} we first formalize the problem and our desired solution.
Then, in \cref{ssec:3dmms_to_the_rescue}, we describe our 3DMM which plays a key role in our approach.
Next, in \cref{ssec:data_driven_transformation_predictor_to_the_rescue}, we explain how to predict the rigid transform between two heads using a neural network.
Finally, we provide details regarding synthesizing training data and training our network in
\cref{ssec:generating_the_training_data,ssec:implementation_details}.


\begin{figure}[tb]
  \centering
  \includegraphics[width=\linewidth]{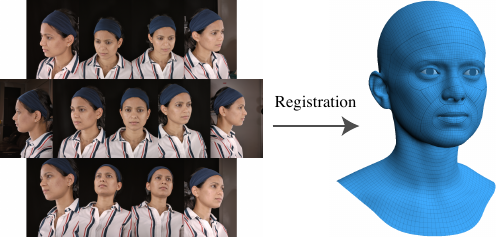}
  \caption{\label{fig:holobooth_example} Our dataset includes $\nframes$ face meshes which were registered to multi-view images. These registrations are used for building our 3DMM and sampling random expressions.}
\end{figure}

\begin{figure*}[tb]
  \centering
  \includegraphics[width=.85\linewidth]{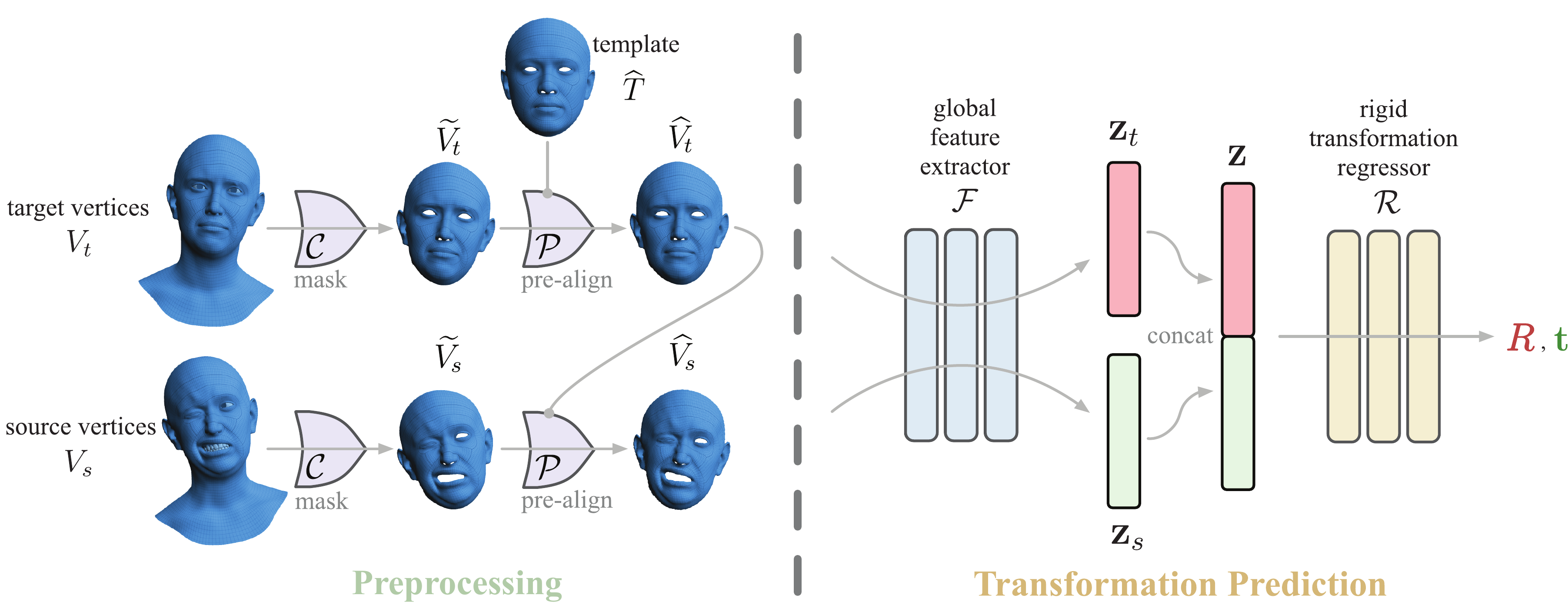}
  \caption{\label{fig:rigid_transformation_predictor} The architecture of the rigid transformation predictor. The network takes a pair $(\vertsskini{s}, \vertsskini{t})$ of source and target skin vertices of the same subject on the input and predicts the rotation $\rotation$ and translation $\translation$ which best aligns the input pair on the output.
  }
\end{figure*}

\subsection{Problem Formulation} \label{ssec:problem_fomulation}
Our goal is to remove unwanted rigid transformation between any two head meshes from a single subject so that the underlying skulls are aligned.

Formally, let $\head = \left(\vertsskin, \vertsskull\right)$ be a head consisting of $\nvertsskin$ observable exterior points $\vertsskin \in \real^{4 \times \nvertsskin}$ and $\nvertsskull$ unobservable skull points $\vertsskull \in \real^{4 \times \nvertsskull}$ in homogeneous coordinates.
Given two misaligned heads $\headi{s} = (\vertsskini{s}, \vertsskulli{s})$ and $\headi{t} = (\vertsskini{t}, \vertsskulli{t})$, representing a \emph{source} and a \emph{target} facial expression of the same subject, our goal is to find a rigid transformation $\matstabstar \in \real^{4 \times 4}$ which best aligns the source and target skulls, i.e. $\matstabstar = \argmin_{\matstab} \energystabcall{s}{t}{\matstab}$, where
\begin{align} \label{eq:stabilization_energy}
    \energystabcall{s}{t}{\matstab} = \frobenius{\matstab \vertsskulli{s} - \vertsskulli{t}}.
\end{align}
Note, that in the ideal noiseless scenario, the skulls align perfectly and $\energystabcall{s}{t}{\matstabstar} = 0$.

Since we only observe the facial exterior points $\vertsskini{i}$ rather than the skull points $\vertsskulli{i}$, we cannot directly minimize the energy $\energystab$ of ~\cref{eq:stabilization_energy}. Instead, our aim is to find a function 
\begin{align}
    {\funcstabcall{s}{t}: \real^{2 \times 4 \times \nvertsskin} \rightarrow \real^{4 \times 4}},
\end{align}
which infers the transformation $\matstabstar$ by only considering observable vertices $\vertsskini{s}, \vertsskini{t}$.

We propose modeling $\funcstab$ as a neural network trained in a supervised way to minimize the energy $\energystab$ of ~\cref{eq:stabilization_energy} for any pair of exterior vertices $(\vertsskini{s}, \vertsskini{t})$.
However, for an observed $\vertsskini{i}$, the mapping $\vertsskini{i} \rightarrow \vertsskulli{i}$ is generally unknown and thus the energy $\energystab$ cannot be readily evaluated during the training.

To address this problem, we propose using a 3DMM to sidestep the missing link between the facial exterior and the skull $\vertsskin, \vertsskull$, thus evaluating the energy $\energystab$ of ~\cref{eq:stabilization_energy} even when $\vertsskull$ is unknown. The details are described in the following sections.

\subsection{3DMMs to the Rescue} \label{ssec:3dmms_to_the_rescue}
\begin{figure}[htb]
  \centering
  \includegraphics[width=.99\linewidth]{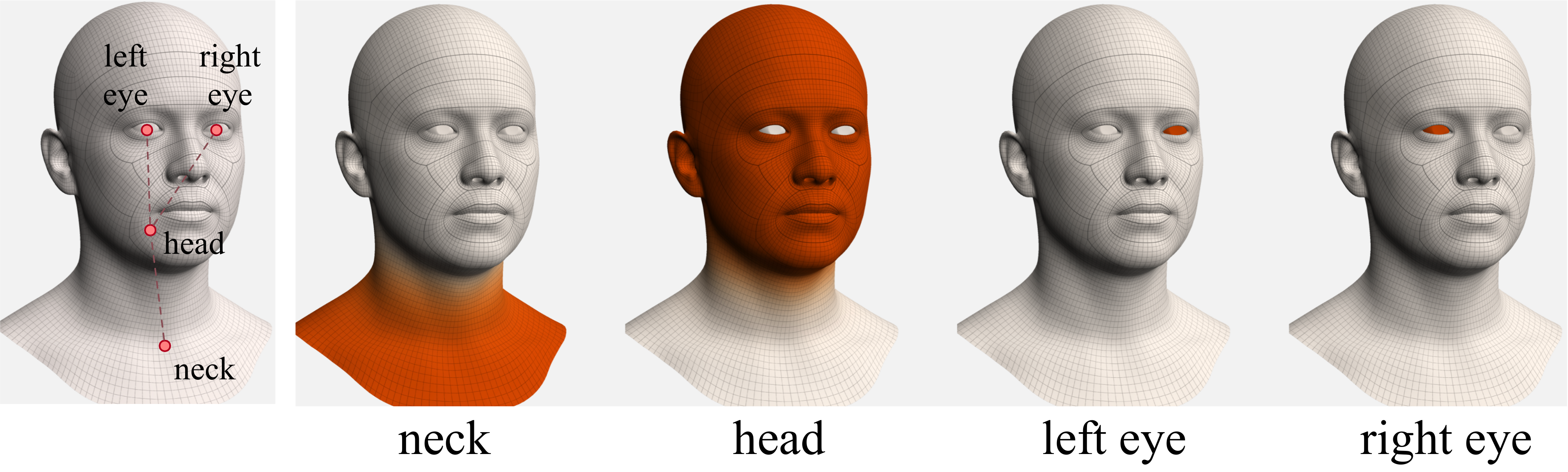}
  \caption{\label{fig:skinning_weights} Schematic view of the kinematic chain of our 3DMM (left) and skinning weights corresponding to the joints (right).}
\end{figure}

Following the literature \cite{li17,paysan09,wood21}, we formulate our 3DMM as a function
$\threedmmwithdata(\threedmmparams): \real^{|\threedmmparams|} \rightarrow \real^{4 \times \nvertsskin}$
which, given model data $\threedmmdata$, takes parameters $\threedmmparams$ and generates $\nvertsskin\!=\!\nthreedmmskinverts$ vertices in homogeneous coordinates.
Our 3DMM model is that of \cite{wood21} with the following differences: (i) we define a custom topology, (ii) we employ 3rd party expression blendshapes \cite{polywink}, and (iii) we use custom artist-painted skinning weights, see \cref{fig:skinning_weights} and \cref{parag:model_data} for more details. Like~\cite{wood21}, our 3DMM includes the eyeballs and teeth.

Model parameters ${\threedmmparams = \left(\paramsid, \paramsexpr, \paramsrot, \paramstransl\right)}$ include identity $\paramsid \in \real^{\nparamsid}$, expression $\paramsexpr \in \real^{\nparamsexpr}$, rotations of $\njoints\!=\!4$ joints $\paramsrot \in \real^{\nparamsrot}$ and global translation $\paramstransl \in \real^{\nparamstransl}$.
Model data $\threedmmdata=\left(\threedmmtempl, \threedmmtempljoints, \basesid, \basesexpr, \basesidjoints, \lbsweights, \jointparents\right)$
includes the
template face vertices $\threedmmtempl \in \real^{4 \times \nvertsskin}$,
template joint locations $\threedmmtempljoints \in \real^{4 \times \njoints}$,
linear vertex identity basis $\basesid \in \real^{|\paramsid| \times 4 \times \nvertsskin}$,
linear expression basis $\basesexpr \in \real^{|\paramsexpr| \times 4 \times \nvertsskin}$,
linear joint identity basis $\basesidjoints \in \real^{|\paramsid| \times 4 \times \njoints}$,
skinning weights $\lbsweights \in \real^{\njoints \times \nvertsskin}$, and
skeletal hierarchy $\jointparents \in \integers^{\njoints}$.

Formally, vertices are generated as follows:
$$
\threedmmwithdata(\threedmmparams) = \linearblendskinning \left( \vertsskinbind, \skinningtransforms, \lbsweights \right)
$$
$\linearblendskinning$ is a standard Linear Blend Skinning function that transforms bind-pose vertices $\vertsskinbind$ by skinning transforms $\skinningtransforms$, with weights $\lbsweights$ controlling how each vertex is affected by each joint.
$$
\skinningtransforms = \mathcal{X} \left(\jointsbind, \paramsrot, \paramstransl; \jointparents \right)
$$
$\mathcal{X}$ builds skinning transforms by propagating per-joint rotations $\paramsrot$ down the kinematic tree defined by $\jointparents$, taking bind-pose joint locations $\jointsbind$ and root joint translation $\paramstransl$ into account.
Vertices and joints in the bind pose are determined using linear bases:
$$
\vertsskinbind = \threedmmtempl + \textstyle{\sum_{i=1}^{\nparamsid}} \paramsid_{i}\basesid_{i} + \sum_{i=1}^{\nparamsexpr}\paramsexpr_{i}\basesexpr_{i} \quad \text{and} \quad
\jointsbind = \threedmmtempljoints + \sum_{i=1}^{\nparamsid} \paramsid_{i}\basesidjoints_{i}
$$

\begin{figure*}[htb]
    \centering
    \hfill
    \begin{subfigure}[b]{0.47\textwidth}
        \centering
        \includegraphics[width=\textwidth]{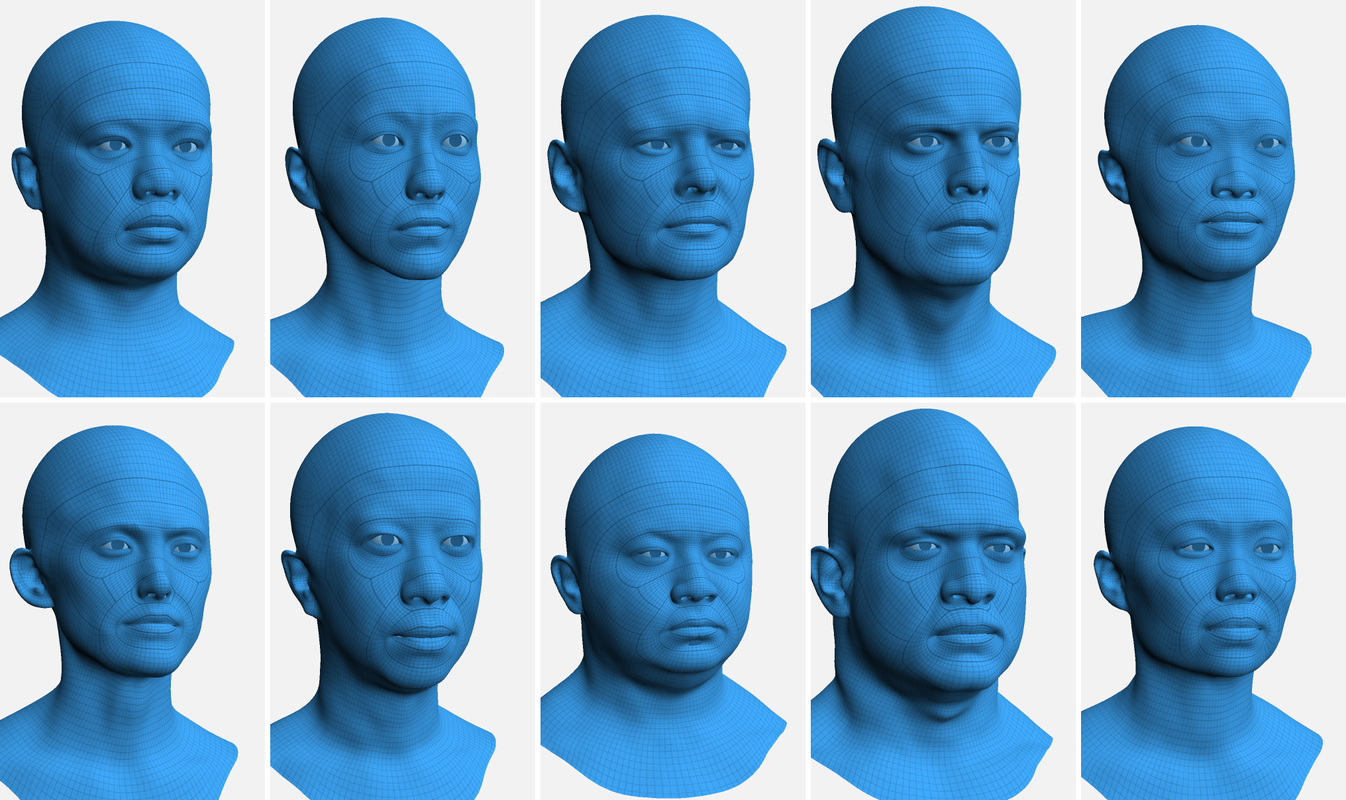}
        \caption{Random identity samples drawn from our generative identity model.}
        \label{fig:dataset_samples_identity}
    \end{subfigure}
    \hfill
    \begin{subfigure}[b]{0.47\textwidth}
        \centering
        \includegraphics[width=\textwidth]{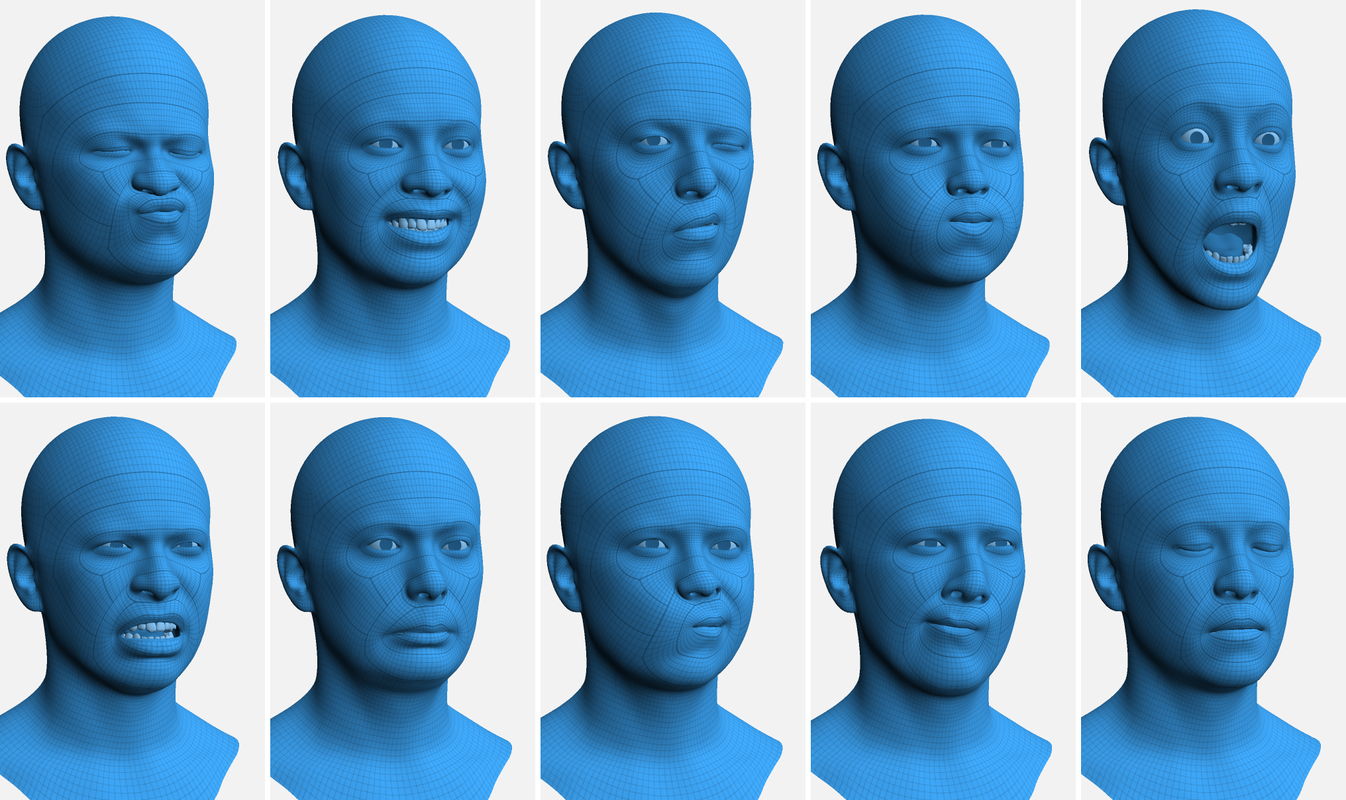}
        \caption{Random expression samples drawn from our dataset of registered faces.}
        \label{fig:dataset_samples_expression}
    \end{subfigure}
    \hfill
    \caption{Our stabilization neural network is trained with synthetic data only. We synthesize realistic and diverse faces by mixing random identities (\cref{fig:dataset_samples_identity}) with random expressions (\cref{fig:dataset_samples_expression}).}
    \label{fig:dataset_samples}
\end{figure*}

\label{parag:model_data}
\paragraph*{Data provenance.}
Our vertex identity basis $\basesid$ is computed by performing PCA on a dataset of registered 3D heads obtained using a multi-camera capture studio. We use a custom optimization-based registration pipeline with standard steps of detecting facial landmarks, enforcing photometric consistency and regularizing surface deformation as done before \cite{fyffe17,li17,cao18}.
The dataset contains $\nframes$ frames of $\nsubjects$ subjects performing a variety of different expressions.
Each registration contains a mesh with $\nvertsskin$ vertices and estimated 3DMM parameters $\threedmmparams$.
See \cref{fig:holobooth_example} for an example.
Our expression basis $\basesexpr$ is FACS-like \cite{ekman78} and was authored by an artist, specifically, we purchased a set of blendshapes Polywink \cite{polywink}. Each basis controls a localized area and guarantees a stable skull. Template joint positions $\threedmmtempljoints$ and skinning weights $\lbsweights$, too, were designed by an artist, see \cref{fig:skinning_weights}.
Joint identity basis $\basesidjoints$ is computed as an average of the vertices corresponding to $\basesid$ weighted by $\lbsweights$.

\paragraph*{3DMMs have a stable skull.}
Generally, 3DMMs do not explicitly model the skull.
But, by design, changing expression parameters $\paramsexpr$ should only deform the exterior vertices $\vertsskin$,
without affecting the hypothetical underlying skull $\vertsskull$.
Thus, for any pair of source and target expressions where
$\vertsskini{s} = \threedmmcallexpri{s}$ and $\vertsskini{t} = \threedmmcallexpri{t}$,
the corresponding skulls should align, i.e. $\vertsskulli{s} = \vertsskulli{t}$.

Suppose we rigidly transform the target skin vertices $\vertsskini{t}$ by any rigid transformation $\matrigidtf$, the skull follows and $\vertsskulli{t} = \matrigidtf\vertsskulli{s}$. By plugging this in the ~\cref{eq:stabilization_energy} we get
\begin{align}
    \energystabcall{s}{t}{\matstab} = \frobenius{\matstab\vertsskulli{s} - \matrigidtf\vertsskulli{s}} = \frobenius{(\matstab - \matrigidtf)\vertsskulli{s}},
\end{align}
and thus $\matrigidtf = \argmin_{S} \energystabcall{s}{t}{\matstab}$. 

In other words, our 3DMM lets us generate facial exterior vertices $\vertsskin$ of various expressions and arbitrary rigid transformations $\matrigidtf$ which can be used to evaluate the energy $\energystab$ even without having access to the actual skull  $\vertsskull$.
We use this convenient property to generate expressions $\vertsskini{i}$ perturbed by rigid transformations $\matrigidtfi{i}$ (see \cref{ssec:data_driven_transformation_predictor_to_the_rescue}), and train a neural network to predict $\matrigidtfi{i}$ which, as we showed above, is the sought-after minimum of energy $\energystab$ of ~\cref{eq:stabilization_energy}.

\subsection{Data-driven Transformation Predictor} \label{ssec:data_driven_transformation_predictor_to_the_rescue}

We model the function $\funcstab$ as a neural network, which takes a pair of misaligned face vertices and produces a rigid transformation $\matstab$. 

\paragraph*{Preprocessing.}\label{parag:preprocessing}
To make the problem easier for $\funcstab$ we preprocess input meshes by
(i) removing less-relevant parts of the mesh that may distract the network (see \cref{parag:head_coverage}), and
(ii) pre-aligning each mesh using a naive method.
Our preprocessing function $(\vertsskini{s}, \vertsskini{t}) \rightarrow (\vertsproctempli{s}, \vertsproctempli{t})$ is defined as follows.
Let $\funcmaskfront: \real^{4 \times \nvertsskin} \rightarrow \real^{4 \times \nvertsfront}$ be a mask extracting $\nvertsfront < \nvertsskin$ vertices corresponding to the frontal face area (see \cref{fig:rigid_transformation_predictor}), $\procrustes(X, Y)$ a Procrustes alignment of points $X$ to $Y$, and $\threedmmtemplfront \in \real^{4 \times \nvertsfront}$ a centered and axis aligned neutral face mesh. We obtain the $(\vertsproctempli{s}, \vertsproctempli{t})$ as (see also \cref{fig:rigid_transformation_predictor}):
\begin{align*}
    \vertsfronti{j} &= \funccallone{\funcmaskfront}{\vertsskini{j}}, j \in \{s, t\} \\
    \vertsproctempli{t} &= \funccalltwo{\procrustes}{\vertsfronti{t}}{\threedmmtemplfront},  \\
    \vertsproctempli{s} &= \funccalltwo{\procrustes}{\vertsfronti{s}}{\vertsproctempli{t}}. 
\end{align*}
Although our 3DMM contains teeth, we do not include them in masked faces $\vertsfronti{i}$.
This is because the teeth are rarely observed in practice, so we cannot rely on them being accurately registered. After extracting the frontal face region via $\funcmaskfront$, the meshes $\vertsfronti{i}$ have $6\,663$ vertices.

\paragraph*{Transformation predictor.}
We model the function $\funcstab$ as neural network with trainable parameters $\nnparams$ as
\begin{align}
    \nnfuncstab(\vertsskini{s}, \vertsskini{t}) = \tfregressor\left(\featureextractor(\vertsskini{s}) \oplus \featureextractor(\vertsskini{t})\right),
\end{align}
where $\featureextractor: \real^{4\nvertsskin} \rightarrow \real^{\latentdim}$ is a global feature extractor mapping flattened skin vertices $\vertsskini{i}$ into a $\latentdim$-dimensional latent code $\latenti{i}$, the operator $\oplus$ represents vector concatenation and $\tfregressor: \real^{2\latentdim} \rightarrow \real^{4 \times 4}$ is a rigid transformation predictor producing the transformation $\matstab$ consisting of rotation $\rotation$ and translation $\translation$ best aligning the input pair $(\vertsskini{s}, \vertsskini{t})$. See~\cref{fig:rigid_transformation_predictor} for the architecture overview. In practice, both $\featureextractor$ and $\tfregressor$ are modeled as multi-layer perceptrons (MLP) accepting a flattened vector of mesh vertices and the combined latent code respectively.

\paragraph*{Synthesizing training data.} \label{ssec:generating_the_training_data}
We synthesize training data for $\nnfuncstab$ using our 3DMM.
Each training sample is a random pair of faces $(\vertsproctempli{s}, \vertsproctempli{t})_{i}$ along with the corresponding ground-truth (GT) transformation $\gttf_{i}$.
Generating realistic and diverse random samples using parametric models is a long-standing open problem, as supported by the body of relevant literature \cite{bogo16,pavlakos19,zanfir20,davydov22,tiwari22}.

Our strategy is simple.
We start with a training dataset of $N$ registered meshes and their estimated 3DMM parameters $\datasetid = \{\paramsid_{i}|1 \leq i \leq N\}$, $\datasetexpr = \{\paramsexpr_{i}|1 \leq i \leq N\}$.
We then fit a normal distribution $\normaldistrib{\meanid}{\diagonal{\stdid}}$ to $\datasetid$.
For a single training pair, we draw one random identity vector from the identity distribution (see \cref{fig:dataset_samples_identity}),
and draw two random expression vectors directly from $\datasetexpr$ (see \cref{fig:dataset_samples_expression}).
We slightly perturb the expression vectors with a small amount of noise $\normaldistrib{\zeros}{\diagonal{\noiseexpr \ones}}$, and pose the meshes with $\threedmm$.
Sampled meshes are preprocessed the same way as shown in \cref{fig:rigid_transformation_predictor} but each Procrustes pre-alignment $\procrustes$ is complemented by a small random rigid transformation to mimic the noise encountered when aligning real-world meshes.

Let $\procrustesmat(X, Y) \rightarrow \matfourbyfour$ compute a rigid Procrustes alignment of points $X$ to $Y$ (i.e. $\procrustesmat$ is equivalent to $\procrustes$ but retrieves the transformation), and let $\funcsamplerigidtf(\noiserot, \noisetransl)$ sample a random rigid transformation given limits $\noiserot, \noisetransl \in \real$ (see Appendix). The process of generating one data sample $(\vertsproctempli{s}, \vertsproctempli{t}, \gttf)$ is described in \cref{alg:data_sampling}.

One might worry: is synthetic data good enough for $\nnfuncstab$ to generalize to real-world registered meshes?
What about the dreaded domain gap?
Fortunately, our results in \cref{ssec:results} indicate
that our dataset synthesis scheme works, and $\nnfuncstab$ does indeed generalize to real-world data.

\SetKwComment{Comment}{//}{}
\begin{algorithm}
\caption{Generating one training sample.}\label{alg:data_sampling}
\SetKwInOut{Input}{Input}
\SetKwInOut{Output}{Output}
\Input{$\threedmm$, $\threedmmtemplfront$, $\meanid$, $\stdid$, $\datasetexpr$, $\noiseexpr$, $\noiserot$, $\noisetransl$}
\Output{$\vertsproctempli{s}, \vertsproctempli{t}, \gttf$}
\Comment{Sample identity, expression parameters and tf. noise.}
$\paramsid \sim \uniformdistribtwo{\meanid - 3\stdid}{\meanid + 3\stdid}$ \\
$\paramsexpr_{s}, \paramsexpr_{t} \sim \uniformdistribone{\datasetexpr} + \normaldistrib{\zeros}{\diagonal{\noiseexpr \ones}}$ \\
$\randomrigidtfi{s} = \funcsamplerigidtf(\noiserot, \noisetransl)$, $\randomrigidtfi{t} = \funcsamplerigidtf(\noiserot, \noisetransl)$ \\
\Comment{Sample source and target vertices.}
$\vertsfronti{s} = \funccallone{\funcmaskfront}{\funccalltwo{\threedmm}{\paramsid}{\paramsexpr_{s}}}$ \\ 
$\vertsfronti{t} = \funccallone{\funcmaskfront}{\funccalltwo{\threedmm}{\paramsid}{\paramsexpr_{t}}}$ \\
\Comment{Noisily align target to template.}
$\tftargettemplate = \funccalltwo{\procrustesmat}{\vertsfronti{t}}{\threedmmtemplfront}$ \\
$\vertsproctempli{t} = \randomrigidtfi{t} \tftargettemplate \vertsfronti{t}$ \\
\Comment{Noisily align source to target.}
$\tfsourcetarget = \funccalltwo{\procrustesmat}{\vertsfronti{s}}{\vertsproctempli{t}}$ \\
$\vertsproctempli{s} = \randomrigidtfi{s} \tfsourcetarget \vertsfronti{s}$ \\
\Comment{Compute the GT transformation.}
$\gttf = \randomrigidtfi{t} \tftargettemplate \invert{\tfsourcetarget} \invert{\randomrigidtfi{s}}$
\end{algorithm}

\paragraph*{Loss functions.} \label{ssec:loss_functions}
Since we know the GT transformation $\gttf_{i}$ for each source-target pair, we train the transformation predictor in a supervised way with the following loss function:
\begin{align} \label{eq:loss}
    \loss &= \lossrot + \weightlosstransl \losstransl, \\
    \lossrot &= \frobenius{\gtrot - \rotation}, \nonumber \\
    \losstransl &= \ltwo{\gttransl -\translation}, \nonumber
\end{align}
where $\gtrot, \gttransl$ are the GT rotation and translation respectively, and $\weightlosstransl$ is a scalar weighing the two loss terms.

\begin{figure}[tb]
  \centering
  \includegraphics[width=.85\linewidth]{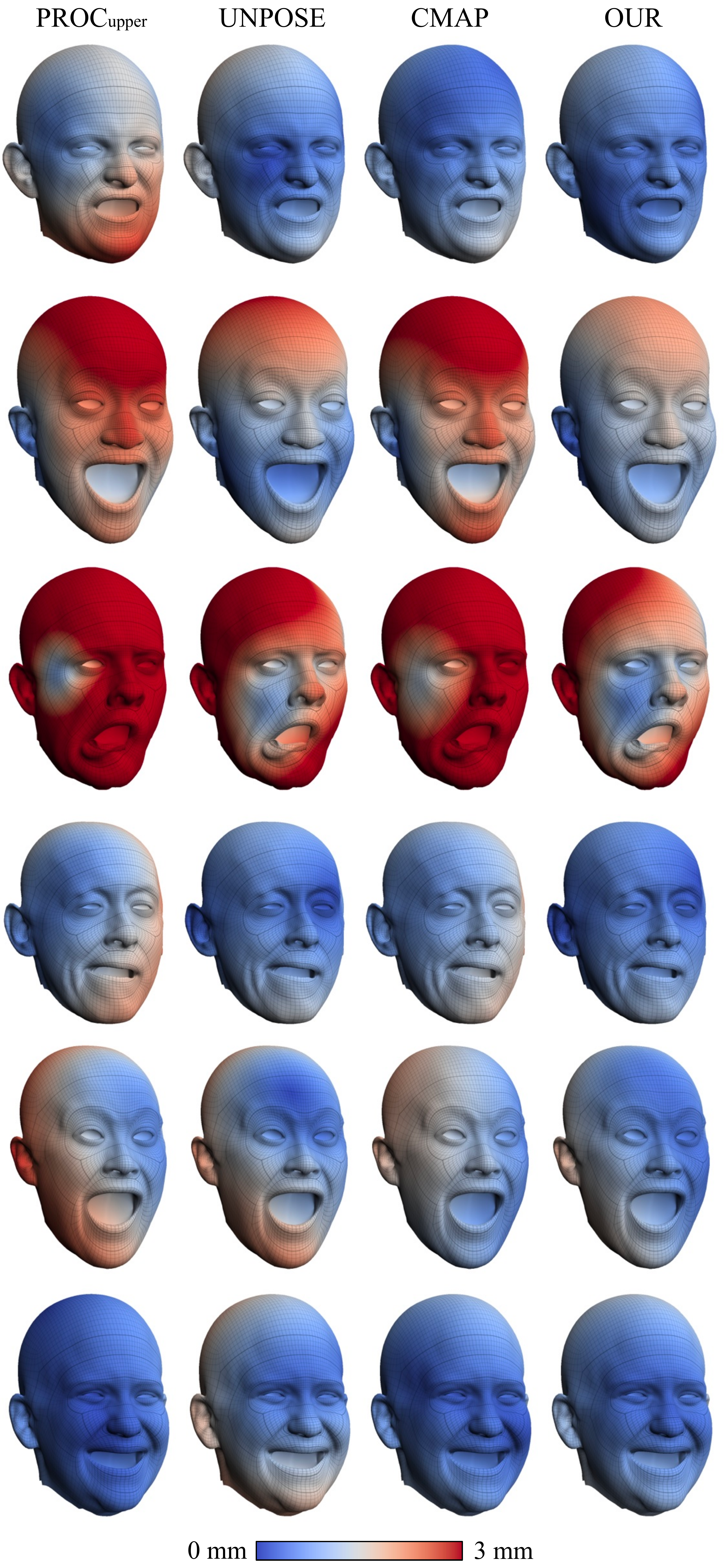}
  \caption{\label{fig:results_error_map} The stabilization error visualized as an error map in the range $[0, 3]$ mm.}
\end{figure}

\begin{figure*}[htb]
  \centering
  \includegraphics[width=\linewidth]{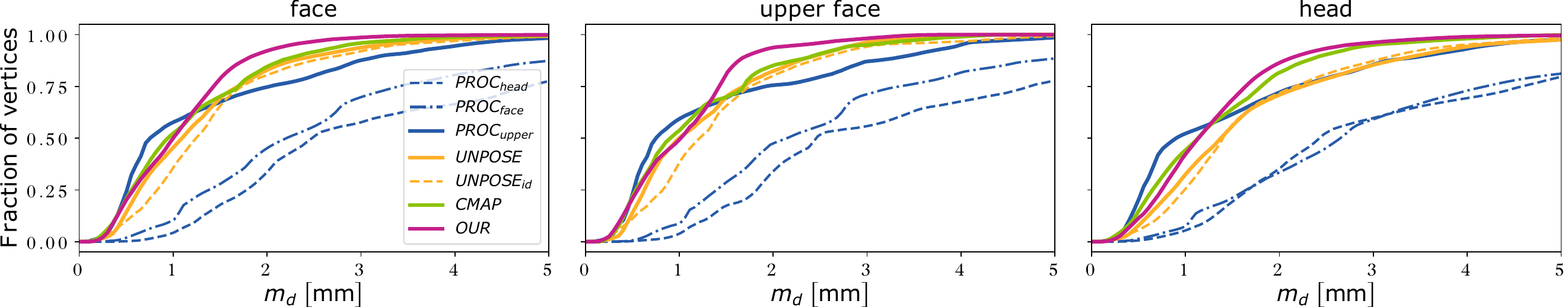}
  \caption{\label{fig:pck} Percentage of correct keypoints (PCK) computed on three different head regions, $\headregion$, $\faceregion$ and $\upperfaceregion$ shown for all the tested methods.}
\end{figure*}

\subsection{Training Details} \label{ssec:implementation_details}
The feature extractor $\featureextractor$ is modeled as a $3$-layered Multilayer Perceptron (MLP) with sizes $(1024, 512, 512)$, the dimension of the latent space $\latentdim = 256$ and the transformation regressor $\tfregressor$ is a $3$-layered MLP with sizes $(512, 512, 512)$. All MLP layers use ReLU except the last linear layer. The network is trained with the Adam optimizer and learning rate $5e^{-5}$ for $125\,000$ iterations. The weight parameter $\weightlosstransl$ of \cref{eq:loss}
is empirically set to $1$ in all our experiments. The output translation $\translation$ is represented as a $3$D vector and the output rotation $\rotation$ as a $6$D representation \cite{zhou19}.

\section{Experiments} \label{sec:experiments}

In the following text we describe the test data and metrics we use to evaluate our method (\our{}), as well as previous methods we compare against.
Furthermore, we present an ablation study and analysis, thus shedding light on the inner workings of our method.

\subsection{Training and Evaluation Data} \label{ssec:training_and_evaluation_data}
We use a real-world dataset of registered meshes introduced in \cref{parag:model_data}. The GT stabilization of the samples is unknown, however, and hard to obtain \cite{beeler14,wu18,cao18}. We follow \cite{beeler14} and manually stabilize $\ngtframes$ expressions across $\ngtsubjects$ randomly selected subjects.

We only select those $\ngtframes$ expressions with visible upper teeth, as it is the only visible head part, which does not deform with changing facial expressions. For one expression per subject, we manually annotate 2D keypoints on the $6$ upper frontal teeth in $5$ camera views and triangulate them to obtain a 3D polyline.

Finally, to stabilize an annotated expression to the remaining ones, an operator manually transforms the source mesh until (i) the 3D teeth polyline projected to the camera view visually aligns with the visible teeth, and (ii) until the two meshes appear visually aligned in a 3D viewer too, see Fig.~\ref{fig:results_teeth} for examples of annotated expressions. We obtain $\ngtstabilizationframes$ source frames with GT transformations to their corresponding target frames.

All the expressions of the $\ngtsubjects$ subjects are removed from the training dataset, and the annotated set is split into $\nvalidationsubjects$ validation and $\ntestsubjects$ test subjects with corresponding $\nvalidationframes$ validation and $\ntestframes$ test expressions.

\subsection{Metrics} \label{ssec:metrics}
Let $\matstab$ and $\gttf$ be the predicted and GT transformation respectively, and $\vertsskin = \predtf \vertsskini{s}$ and $\gtverts = \gttf \vertsskini{s}$ be the predicted and GT stabilized source vertices respectively. Furthermore, let $\ithsamplejthvert{i}{j}$ denote $j$-th vertex of the $i$-th dataset sample, where ${1 \leq j \leq \ntestverts}$ and ${1 \leq i \leq \ntestsamples}$. We use the following metrics to quantitatively evaluate \our{} and the competing methods.

\paragraph*{Mean vertex distance ($\metricmeanvertdist$).} The metric is computed as ${\metricmeanvertdist = \frac{1}{\ntestsamples \ntestverts} \sum_{i=1}^{\ntestsamples} \sum_{j=1}^{\ntestverts} \ltwo{\ithsamplejthvertgt{i}{j} - \ithsamplejthvertpred{i}{j}}}$.

\paragraph*{Maximum vertex distance ($\metricmaxvertdist$).} Following \cite{beeler14}, we complement $\metricmeanvertdist$ with the maximum vertex distance averaged over the $\ntestsamples$ samples: ${\metricmaxvertdist = \frac{1}{\ntestsamples}\sum_{i=1}^{\ntestsamples} \max_{j} \ltwo{\ithsamplejthvertgt{i}{j} - \ithsamplejthvertpred{i}{j}}}$.

\paragraph*{Area Under Curve ($\metricaucparag$).}
We also compute the percentage of correct keypoints (PCK) metric popular in human pose estimation domain \cite{zheng23,chen22,kocabas19}, evaluate it in the range of $[\pckmin, \pckmax]$ mm and report the area under curve (AUC).

In order to gain further insight into which facial parts are the most challenging for the methods to align, we evaluate all the metrics on various head mesh masks as explained in \cref{ssec:results}.

\subsection{Competing Methods} \label{ssec:competing_methods}
Existing approaches are often subject to prohibitive requirements, such as manual keypoint annotation \cite{beeler14,wu16}, temporally consistent sequence of input meshes \cite{lamarre18,cao18} or reconstructed eyeballs \cite{fyffe17}, none of which are needed by \our{}. For fair comparison, we thus select the following methods which have looser restrictions.

\label{parag:procrustes_alignment}
\paragraph*{Procrustes alignment.}
Since registered face meshes are in correspondence, we can use Procrustes \cite{gower75} to rigidly align them \cite{vlasic05}.
To limit the impact of potentially irrelevant facial areas, we evaluate Procrustes on three vertex subsets corresponding to the full \emph{head}, \emph{face} and \emph{upper face}, the last one being suggested in \cite{weise11,bouaziz13}, which we refer to as $\procrusteshead$, $\procrustesface$ and $\procrustesupper$.

\begin{table*}[htbp]
  \centering
  \caption{The comparison of the methods on three different head regions. The metrics $\metricmeanvertdist$ and $\metricmaxvertdist$ are measured in milimeters, $\metricauc$ in percent.}
    \begin{tabular}{lcccrcccrccc}
\cmidrule{1-4}\cmidrule{6-8}\cmidrule{10-12}    \multicolumn{4}{c}{\textbf{face}} &       & \multicolumn{3}{c}{\textbf{upper face}} &       & \multicolumn{3}{c}{\textbf{head}} \\
\cmidrule{1-4}\cmidrule{6-8}\cmidrule{10-12}    \textbf{method} & $\metricmeanvertdist \downarrow$ & $\metricmaxvertdist\downarrow$ & $\metricauc\uparrow$ &       & $\metricmeanvertdist\downarrow$ & $\metricmaxvertdist\downarrow$ & $\metricauc\uparrow$ &       & $\metricmeanvertdist\downarrow$ & $\metricmaxvertdist\downarrow$ & $\metricauc\uparrow$ \\
\cmidrule{1-4}\cmidrule{6-8}\cmidrule{10-12}    $PROC_{head}$ & $3.46\pm2.37$ & $14.76$ & $40.30$ &       & $3.48\pm2.32$ & $11.85$ & $40.15$ &       & $3.34\pm2.32$ & $14.76$ & $42.03$ \\
    $PROC_{face}$ & $2.84\pm2.21$ & $14.22$ & $49.67$ &       & $2.71\pm1.86$ & $10.85$ & $50.08$ &       & $3.39\pm2.51$ & $14.22$ & $42.82$ \\
    $PROC_{upper}$ & $1.40\pm1.28$ & $8.36$ & $72.15$ &       & $1.38\pm1.24$ & $5.92$ & $72.32$ &       & $1.53\pm1.33$ & $8.36$ & $69.63$ \\
    $UNPOSE$ & $1.32\pm0.94$ & $7.75$ & $73.50$ &       & $1.25\pm0.80$ & $4.44$ & $74.73$ &       & $1.72\pm1.21$ & $7.99$ & $65.85$ \\
    $UNPOSE_{id}$ & $1.46\pm0.98$ & $8.08$ & $70.82$ &       & $1.42\pm0.93$ & $5.59$ & $71.40$ &       & $1.71\pm1.10$ & $8.08$ & $65.86$ \\
    $CMAP$ & $1.20\pm0.86$ & $6.15$ & $75.71$ &       & $1.19\pm0.85$ & $4.52$ & $75.96$ &       & $1.33\pm0.89$ & $\mathbf{6.15}$ & $73.14$ \\
    \textbf{OUR} & $\mathbf{1.08\pm0.64}$ & $\mathbf{5.37}$ & $\mathbf{78.03}$ &       & $\mathbf{1.07\pm0.63}$ & $\mathbf{3.85}$ & $7\mathbf{8.39}$ &       & $\mathbf{1.29\pm0.78}$ & $6.36$ & $\mathbf{74.05}$ \\
\cmidrule{1-4}\cmidrule{6-8}\cmidrule{10-12}    \end{tabular}%
  \label{tab:results}%
\end{table*}%

\label{parag:3dmm_unposing}
\paragraph*{3DMM unposing.}
Each registered face comes with estimated 3DMM parameters
so we can stabilize the pair by unposing both the source and the target to the bind pose, as done in \cite{li17}.
Using our 3DMM introduced in \cref{ssec:3dmms_to_the_rescue}, and given source and target parameters $\threedmmparams_{x} = (\paramsid_{x}, \paramsexpr_{x}, \paramsrot_{x}, \paramstransl_{x})$, where $x \in \{s, t\}$, the unposed mesh vertices are obtained as $\threedmm(\paramsid_{x}, \paramsexpr_{x}, \zerorotparams, \zerotranslparams)$ with $\zerorotparams, \zerotranslparams$ being zero rotation and translation parameters. The predicted transformation $\matstab$ is thus fully defined by $\paramsrot_{s}, \paramstransl_{s}, \paramsrot_{t}, \paramstransl_{t}$.
Since only expressions of the same subject are stabilized, we also consider the cases where we refit the 3DMM model to the source and target meshes so that the identity parameters $\paramsid$ are the same, i.e. $\paramsid_{s} = \paramsid_{t}$.
We refer to these two flavours as $\unpose$ and $\unposesharedid$.
Also note that unposing a 3DMM comes with a particular drawback.
When fitting the 3DMM to the observed meshes, the non-zero fitting error is arbitrarily distributed between the global pose and the 3DMM parameters, which precludes perfect stabilization.
Furthermore, the necessary step of fitting the 3DMM to the observed meshes renders the method slower than the other approaches.

\label{parag:learned_confidence_map}
\paragraph*{Learned confidence map.}
We reimplemented the confidence-map-based rigid stabilization module of \cite{wu18}, and trained it on our own data.
We found the original formulation produced unsatisfactory results, on par with the basic $\procrusteshead$.
Therefore, we modified the method to encourage high global contrast and local spatial consistency, and found the best set of hyper-parameters on the validation set, please see the Appendix for more details.
We only experiment with the modified variant and refer to it as \methodconfidencemap{}.

\subsection{Results} \label{ssec:results}

We now compare our method to existing work quantitatively and qualitatively using the annotated test set described in \cref{ssec:training_and_evaluation_data}.
Furthermore, an ablation study motivates our design choices. 

\paragraph*{Quantitative results.} \label{parag:quantitative_results}

We evaluate all methods on the metrics defined in \cref{ssec:metrics}.
To reveal any biases towards specific parts of the head, we measure error across three regions.
\emph{\headregion} discards the neck which is not relevant to stabilization quality,
\emph{\faceregion} considers the frontal face area, and
\emph{\upperfaceregion} considers the forehead and nose only, which is typically the most robust to changing expressions.

As can be seen in \cref{tab:results}, \our{}  yields the best performance improving upon the best performing baseline \methodconfidencemap{} by $10\%$.
Among all the variants of Procrustes alignment, $\procrustesupper$ performs the best, which is in line with the assumptions made by prior work \cite{bouaziz13,weise11}, but still falls short of the remaining methods.
While $\unpose$ improves on the Procrustes alignment, it suffers from the underlying non-zero fitting error, as discussed in \cref{ssec:competing_methods}, manifesting in imprecise stabilization.
Also note, that despite guaranteeing consistent identity parameters, $\unposesharedid$ underperforms $\unpose$ which we attribute to worse 3DMM fitting caused by decreased model flexibility.

\methodconfidencemap{} yields the best result among the prior work.
Remarkably, the method is based on Procrustes alignment, but shows that seemingly rudimentary rigid alignment can yield decent results, if one learns alignment mask from the data.
Despite that, the linear nature of the Procrustes alignment limits the performance when compared to the ML-based solution of \our.

To demonstrate the robustness of the methods, we further show curves evaluating the PCK metric in the range $[0, 5]$ mm in \cref{fig:pck}. It is evident that \our{} generally outperforms the other methods.
Interestingly, $\procrustesupper$ produces stabilization with a much higher fraction of vertices below the error of $1$ mm, but deteriorates above this mark. This phenomenon is due to the fact that $\procrustesupper$ focuses solely on the forehead while ignoring the rest of the face contributing to the high overall error.

\paragraph*{Qualitative results.} \label{parag:qualitative_results}

\begin{figure*}[htb]
  \centering
  \includegraphics[width=.99\linewidth]{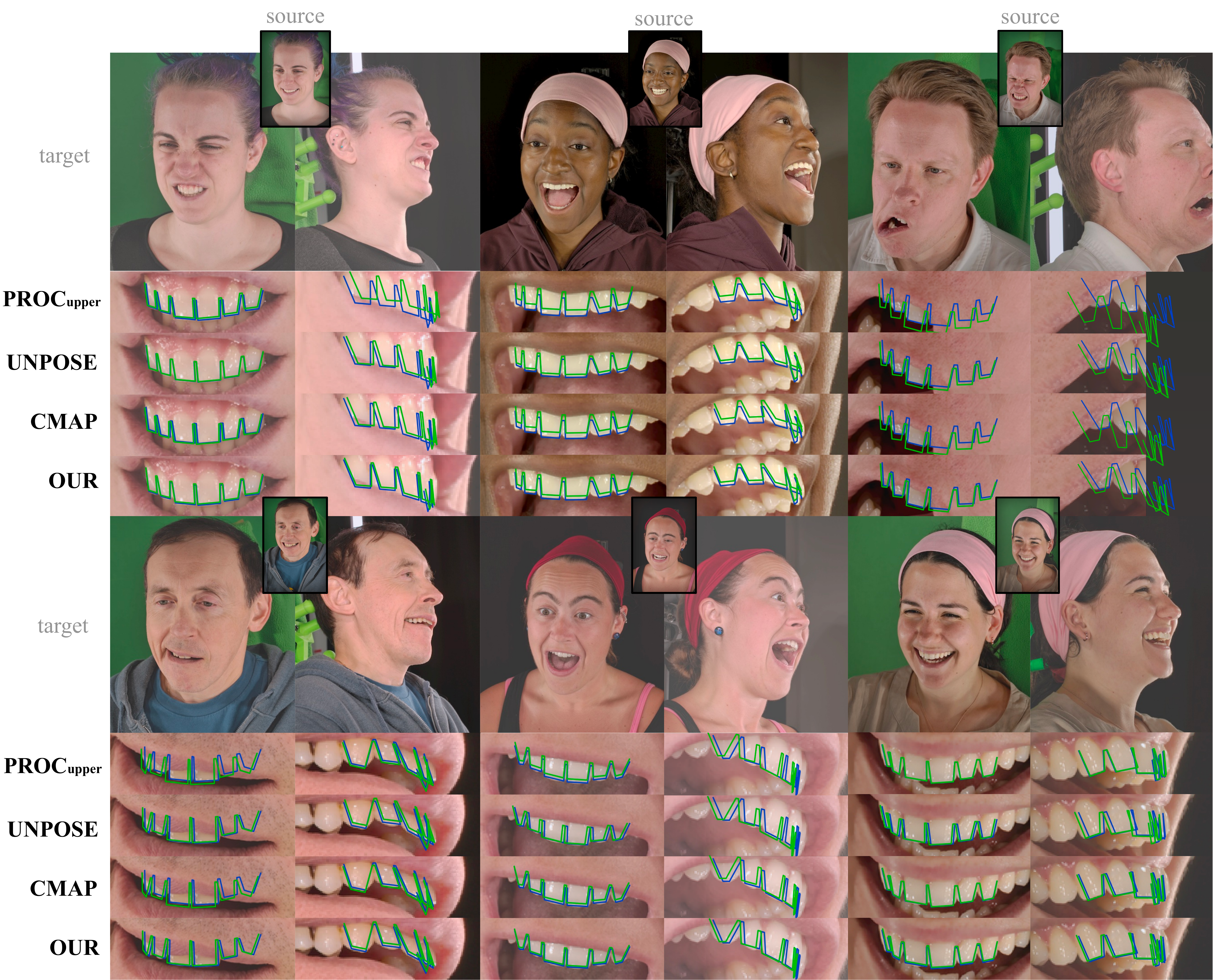}
  \caption{\label{fig:results_teeth} Visual comparison of the methods showing predicted (green) and GT (blue) 3D teeth line projected to the front and side views.}
\end{figure*}

\Cref{fig:results_error_map,fig:results_teeth} contain a visual comparison of the methods on the task of stabilizing arbitrary expression pairs from the test set. In \cref{fig:results_error_map} we show the spatial error of stabilizing an expression pair using GT and predicted transformation. In \cref{fig:results_teeth} we transform the GT 3D teeth poly-line (see \cref{ssec:training_and_evaluation_data}) from a source to a target expression using transformations predicted by each method, project the poly-line to two camera views of the target, and overlay it with the GT one.
We only show the best performing variants of Procrustes ($\procrustesupper$) and 3DMM unposing (\unpose{}).

The qualitative results are in line with the aforementioned observations. $\procrustesupper$ produces the least precise stabilization, while $\unpose$ and \methodconfidencemap{} generally yield high-quality alignment, where the errors start appearing for more extreme and/or asymmetric facial expressions such as the top-right subject moving their jaw sideways or the top left subject snarling in \cref{fig:results_teeth}. \our{}, too, suffers from misalignments on more complex expressions but is generally more robust.

The quality of our results is best assessed in videos, so please refer to our supplementary webpage.

\begin{table}[htbp]
  \centering
  \caption{Choice of the head mesh region on the input to \our{} matters, \theatermask performs the best on the validation set.}
    \begin{tabular}{rccc}
    \toprule
    \multicolumn{1}{c}{\textbf{metrics}} & Full  & Face\&Neck & Superhero \\
    \midrule
    \multicolumn{1}{c}{\boldmath{}\textbf{$\metricmeanvertdist\downarrow$}\unboldmath{}} & $1.29\pm0.89$ & $\mathbf{1.15\pm0.77}$ & $1.23\pm0.88$ \\
    \multicolumn{1}{c}{\boldmath{}\textbf{$\metricmaxvertdist\downarrow$}\unboldmath{}} & $5.90$ & $4.85$ & $\mathbf{4.47}$ \\
    \multicolumn{1}{c}{\boldmath{}\textbf{$\metricauc\uparrow$}\unboldmath{}} & $73.98$ & $\mathbf{76.76}$ & $75.05$ \\
    \midrule
          & \includegraphics[width=.23\linewidth]{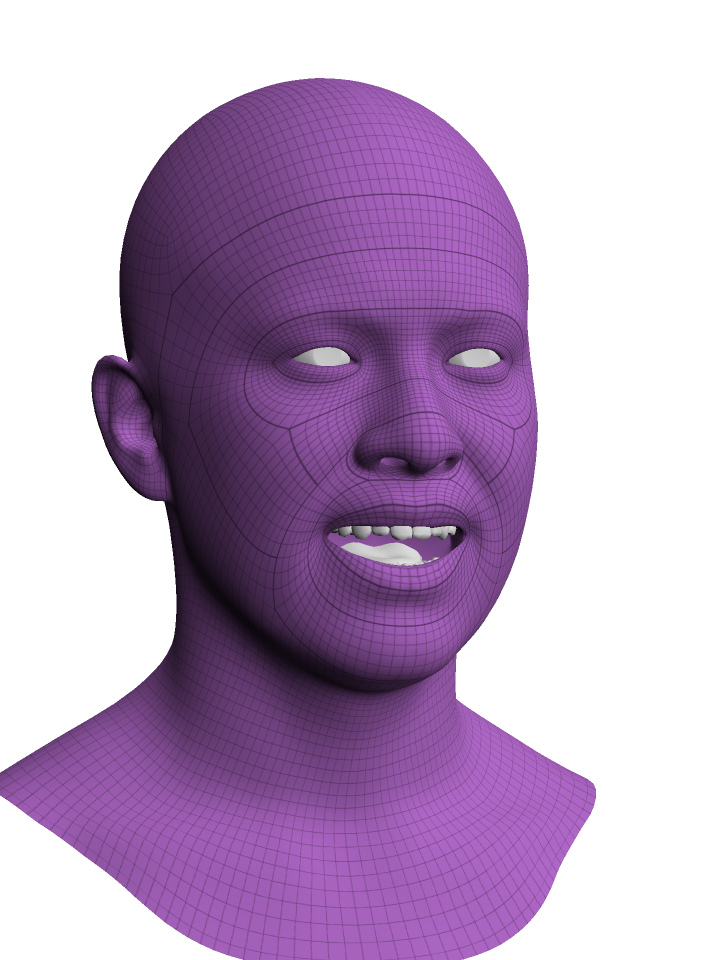} & \includegraphics[width=.23\linewidth]{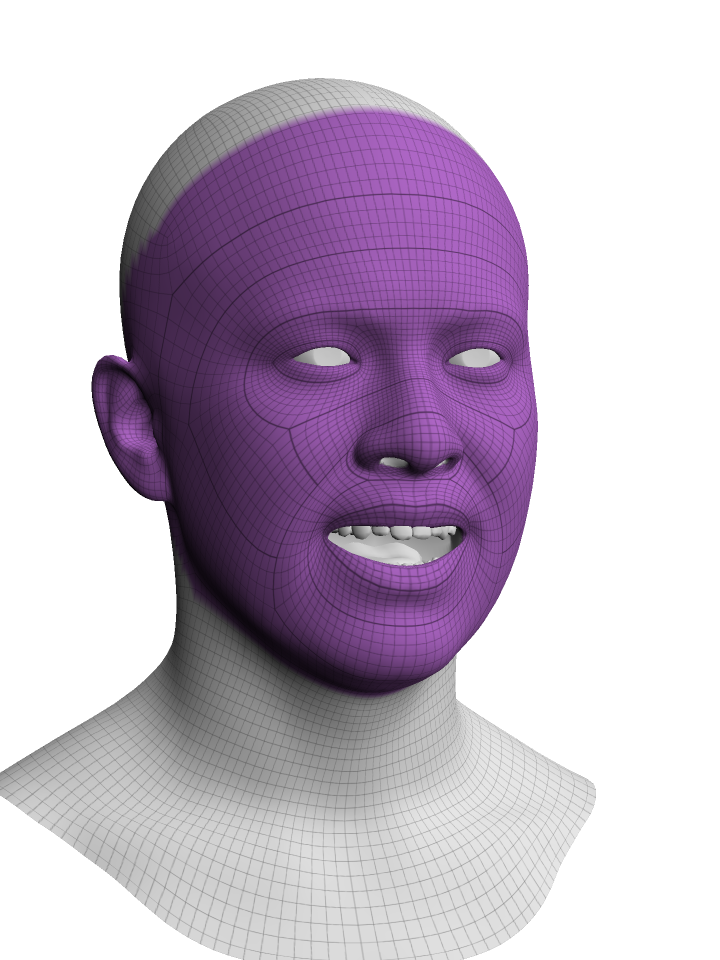} & \includegraphics[width=.23\linewidth]{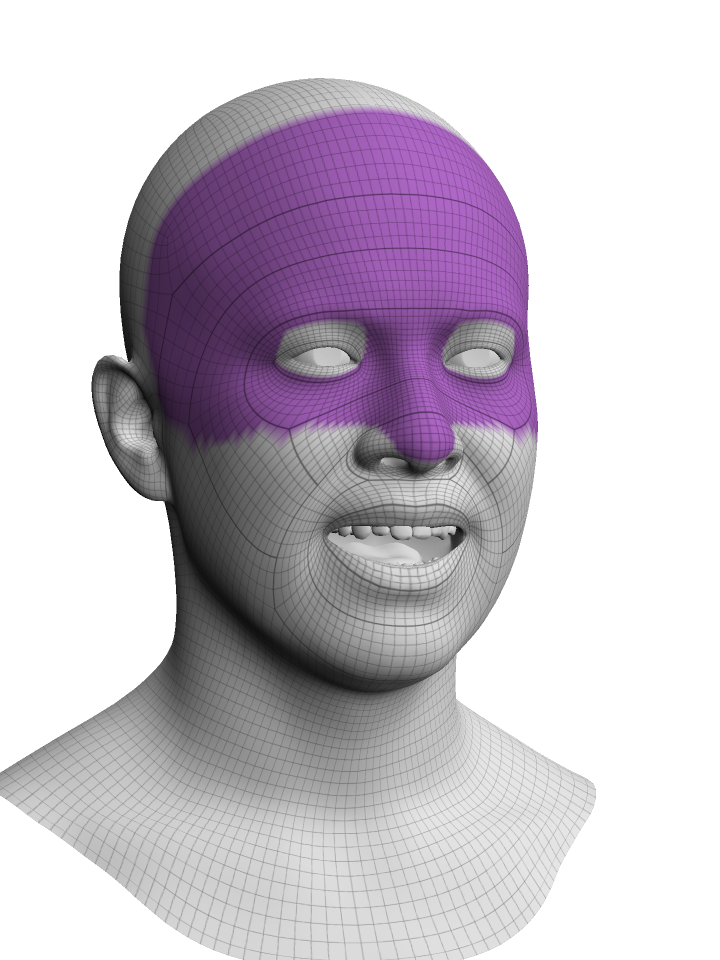} \\
    \end{tabular}%
  \label{tab:ablation_head_coverage}%
  \vspace{-0.5cm}
\end{table}%

\subsection{Ablation Study and Method Analysis} \label{ssec:ablation_study}

Here we discuss the choices taken to design and train the model.

\paragraph*{Head coverage.} \label{parag:head_coverage}

The registered head meshes in our dataset contain the full head and neck. However, the goal is to align the underlying skulls and furthermore, as shown in the literature \cite{weise11,bouaziz13}, not all parts of the face are equally relevant for stabilization. Theoretically, an MLP-based model, which takes flattened meshes on the input, should be able to learn to ignore the irrelevant parts. However, we experimentally show this not to be the case.

We choose three regions of the input meshes shown in \cref{tab:ablation_head_coverage} and train \our{} only on the corresponding vertices. The result of the evaluation on the validation dataset is summarized in \cref{tab:ablation_head_coverage}. It is clear that both feeding the model irrelevant mesh regions (neck and back of the head of $\fullmask$), and pruning the parts that potentially carry a useful signal (cheeks and jaw of $\superheromask$) are detrimental to the performance.

\paragraph*{Training dataset size.} \label{parag:size_of_the_training_dataset}

\begin{figure}[htb]
  \centering
  \includegraphics[width=.95\linewidth]{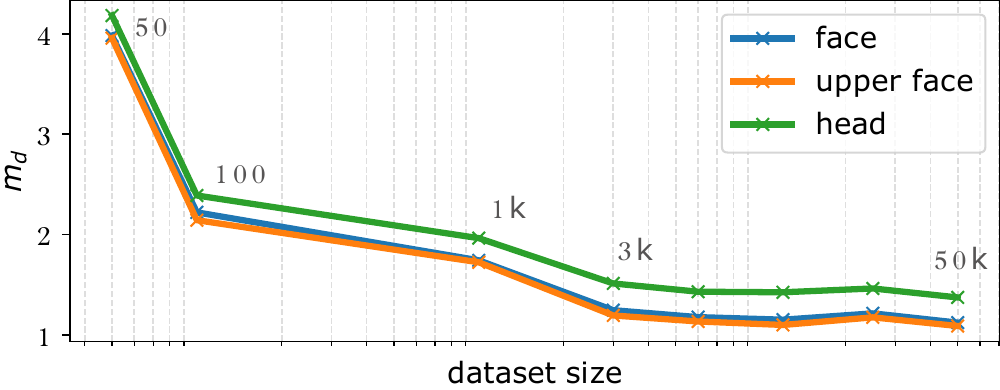}
  \caption{\label{fig:ablation_dataset_size}Impact of the size of the training set on the performance of our method.}
\end{figure}

As discussed in \cref{ssec:generating_the_training_data}, a 3DMM allows us to generate a dataset of unlimited size. How big a dataset is necessary to reach a good performance? We trained \our{} on the randomly generated datasets of various sizes and evaluated them on the validation dataset, the results are presented in \cref{fig:ablation_dataset_size}. It can be seen that training our model on datasets larger than $~3000$ samples yields diminishing returns. While \our{} was trained on a dynamically generated (thus de facto infinite) dataset, in scenarios with limited computation resource, much smaller datasets will suffice.

\section{Conclusion} \label{sec:conclusion}

We presented a novel learning-based approach for rigidly stabilizing face meshes with arbitrary expressions.
Synthetic data played a key part.
We designed a simple but effective scheme for synthesizing training pairs of misaligned expressive faces using a 3DMM.
We used the resulting dataset to train a neural network that directly predicts the rigid transform between any two input meshes so the underlying skulls are aligned.

Our method does not require the input meshes to be temporally consistent.
That is, any pair of arbitrarily differing expressions can be stabilized. This makes our approach generally useful for practitioners seeking to stabilize a continuous facial performance or random expression sets alike, where the typical downstream applications include character deformation transfer or building a custom human head parametric model, where spurious global transformations degrade the expressive capacity of the model.

As our method operates on independent mesh pairs, it can be heavily parallelized allowing for fast processing of large datasets. For example, a performance of $1\,000$ frames is stabilized in $\sim6$ seconds on an Nvidia A100 GPU.
Finally, we show through quantitative and qualitative experiments that our approach outperforms prior work.

Limitations remain.
First, our method relies on sampling a 3DMM, which first needs to be built.
However, such a task is well understood as evidenced by a rich body of literature \cite{egger20,li17,paysan09}.
Second, the method operates on registered meshes with a common topology and thus
it is intended to be used with production studio capture pipelines rather than in-the-wild raw 3D scans for which the method would not work out-of-the-box. Finally, to sample diverse yet realistic faces at training time, our method needs access to a large dataset of 3DMM parameters fitted to scan meshes.


\bibliographystyle{eg-alpha-doi}  
\bibliography{references}


\newpage ~ \newpage

\setcounter{section}{0}
\twocolumn[
  \begin{@twocolumnfalse}
    \vspace{1cm}
    \begin{center}
        \Huge{Appendix}
    \end{center}
    \vspace{1cm}
    
  \end{@twocolumnfalse}
]
We expand on the details behind sampling random rigid transformations in \cref{sec:sampling_random_rigid_transformtions}, and we provide details of training the modified prior work of \cite{wu18} in \cref{sec:learned_confidence_map_implementation}.
\section{Sampling Random Rigid Transformations} \label{sec:sampling_random_rigid_transformtions}

The Algorithm 1 in the main paper describes the process of generating the misaligned pairs of face meshes to train our method. We introduced the function $\funcsamplerigidtf(\noiserot, \noisetransl): \real \times \real \rightarrow \real^{4 \times 4}$ which, given two scalars $\noiserot, \noisetransl$, generates a random rigid transformation
\begin{align*}
    \rigidtf = 
        \begin{bmatrix}
            \rotation & \translation \\
            \transpose{\zerovector} & 1
        \end{bmatrix},
\end{align*}
where $\rotation \in \real^{3 \times 3}$ is a rotation matrix and $\translation \in \real^{3}$ is a translation vector.

We generate the rotation matrix $\rotation$ by sampling an angle $\randomangle$ from a normal distribution parameterized by $\noiserot$ and a random axis $\randomaxis \in \real^{3}$ as follows:
\begin{align*}
    \randomangle &\sim \normaldistrib{0}{\noiserot} \\
    x, y, z &\sim \uniformdistribtwo{-1}{1} \\
    \randomaxis &= \transpose{[x, y, z]} \\
    \rigidtf &= \funcangleaxis(\randomangle, \randomaxis),
\end{align*}
where $\funcangleaxis$ converts an angle and an axis to the rotation matrix. Finally, we generate the translation vector $\translation$ as follows:
\begin{align*}
    x, y, z &\sim \normaldistrib{\zerovector}{\noisetransl} \\
   \translation &= \transpose{[x, y, z]}.
\end{align*}

\section{Learned Confidence Map Implementation} \label{sec:learned_confidence_map_implementation}

As discussed in Section 4.3. of the main paper, we re-implemented the existing work of \cite{wu18}, referred to as \methodconfidencemap{}, but found the original formulation to produces unsatisfactory results. This section details the modification and training procedure we applied to boost its performance.

At its core, \methodconfidencemap{} performs a Procrustes alignment. However, the strength of the methods come from the learned facial mask, which determines the facial regions to be considered for the rigid alignment. Let $\sourceverts, \targetverts \in \real^{4 \times \nverts}$ be the source and target vertices in homogeneous coordinates, $\weights \in [0, 1]^{\nverts}$ per-vertex weights, $\procrustes(\sourceverts, \targetverts)$ Procrustes alignment producing the source vertices aligned to the target, and $\hadamard$ Hadamard product. We can compute weighted source and target vertices as 
\begin{figure}[tb]
  \centering
  \includegraphics[width=\linewidth]{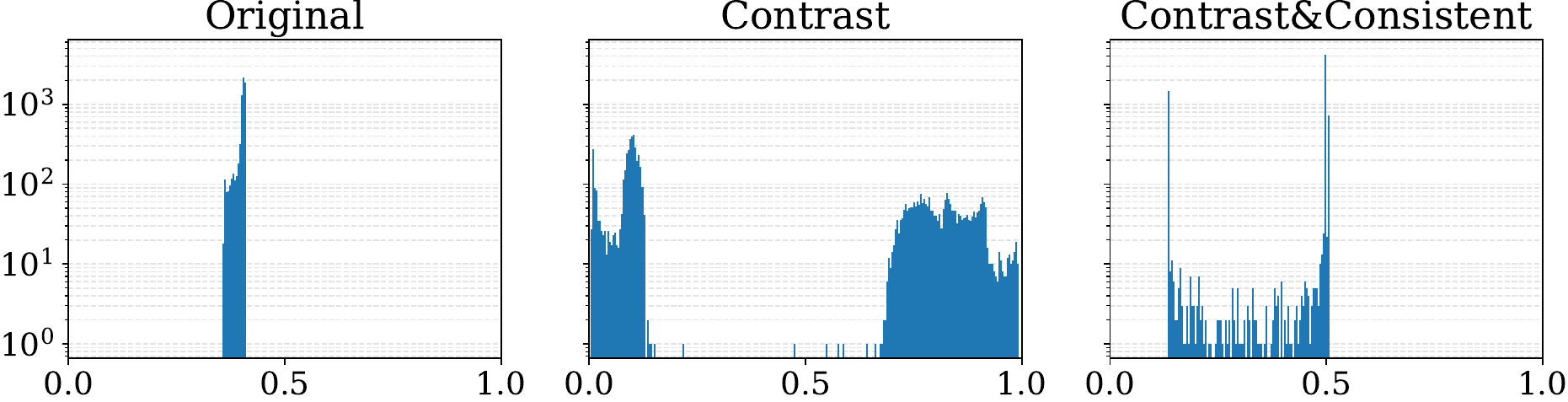}
  \caption{\label{fig:confidence_map_histograms} Distribution of the facial mask weights learned by the various versions of \methodconfidencemap{}.}
\end{figure}
\begin{figure}[tb]
  \centering
  \includegraphics[width=\linewidth]{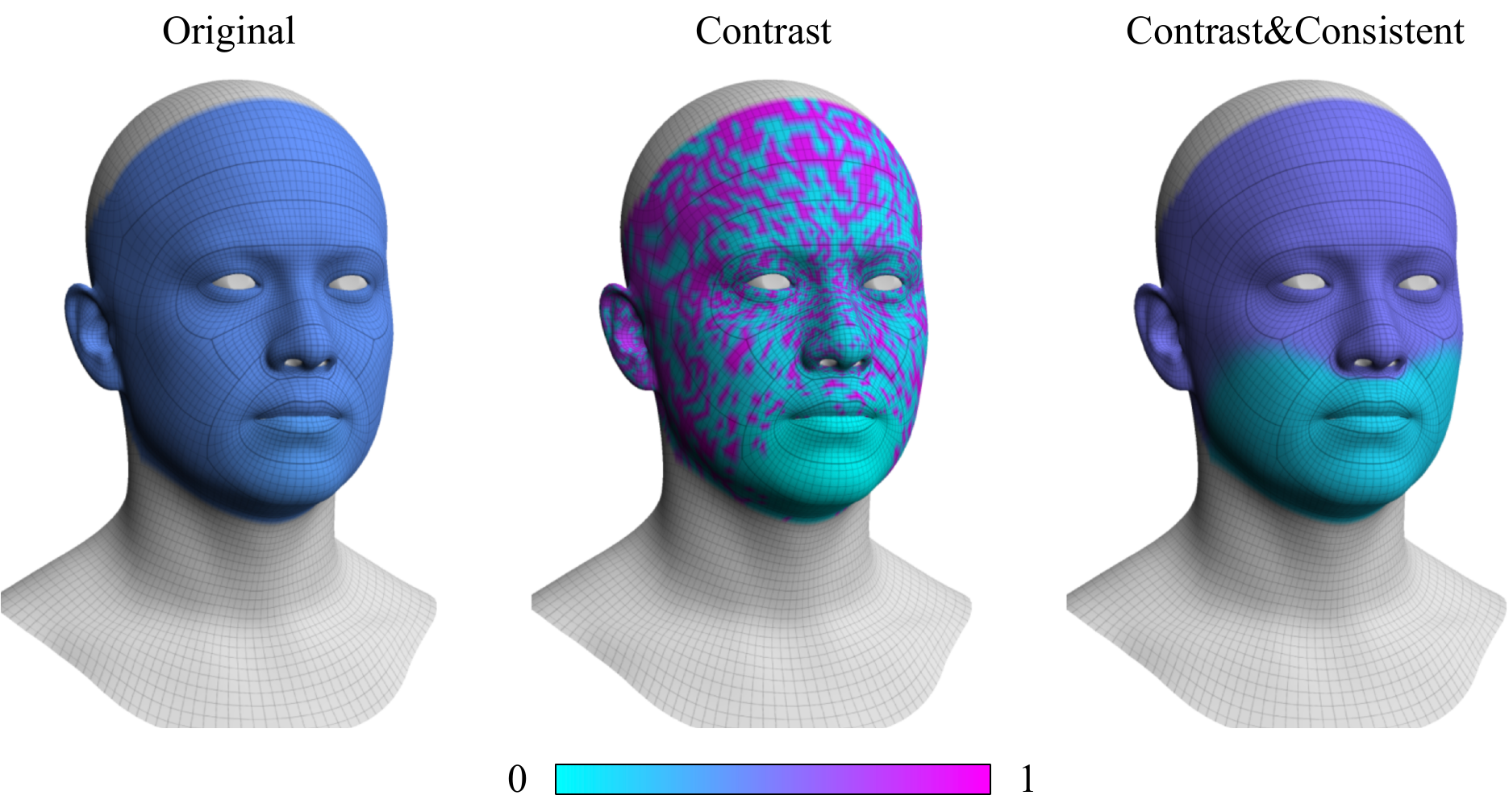}
  \caption{\label{fig:confidence_map_masks} Facial masks learned by the various versions of \methodconfidencemap{}.}
\end{figure}
\begin{align*}
    \maskedsourceverts &= \tiledweights \hadamard \sourceverts \\
    \maskedtargetverts &= \tiledweights \hadamard \targetverts \\
    \tiledweights &= \transpose{\onesvectorfour} \transpose{\weights}.
\end{align*}
Then, \methodconfidencemap{} finds the optimal weights $\weights^{*}$ by solving the following minimization problem:
\begin{align*}
    \weights^{*} &= \argmin_{\weights}\ \alphadata \lossdata + \alphareg \lossreg \\ 
    \lossdata &= \frobenius{\procrustes(\maskedsourceverts, \maskedtargetverts) - \maskedtargetverts}^{2} \\
    \lossreg &= \max{\left(0, \rho \nverts - \ltwo{\weights}^{2}\right)},
\end{align*}
where $\alphadata, \alphareg$ are loss term weights, and $\rho$ is a hyperparameter set by the authors to $0.4$.

We refer to this energy formulation as \cmaporiginal{}, and we found that optimizing this problem leads to a very narrow distribution of weights which do not clearly prefer some facial areas from others, as can be seen in \cref{fig:confidence_map_histograms} and \cref{fig:confidence_map_masks}. This further leads to suboptimal results, as shown in \cref{tab:quantitative_cmap_comparison}.

To encourage higher contrast in the learned weights, we add an additional energy term
\begin{align*}
    \losscontrast = -\funcstd(\weights),
\end{align*}
where $\funcstd$ computes standard deviation over a vector of values. This variant, which we refer to as \cmapcontrast{}, is forced to make a clear decision about which facial areas are relevant for the rigid alignment, as seen in \cref{fig:confidence_map_histograms} and \cref{fig:confidence_map_masks}. It is evident that the method tends to discard the jaw area, which is typically the least stable part across an expression set. While the results improve, as seen in \cref{tab:quantitative_cmap_comparison}, the mask appears noisy which harms the performance.

\begin{table}[tbp]
  \centering
  \caption{Quantitative comparison of the \methodconfidencemap{} variants.}
    \begin{tabular}{lccc}
    \toprule
    \textbf{region} & \boldmath{}\textbf{$\metricmeanvertdist\downarrow$}\unboldmath{} & \boldmath{}\textbf{$\metricmaxvertdist\downarrow$}\unboldmath{} & \boldmath{}\textbf{$\metricauc\uparrow$}\unboldmath{} \\
    \midrule
    Original & $2.37\pm1.51$ & $9.50$ & $54.94$ \\
    Contrast & $1.72\pm1.05$ & $6.21$ & $65.57$ \\
    Contrast\&Consistent & $\mathbf{1.51\pm1.00}$ & $\mathbf{5.43}$ & $\mathbf{69.66}$ \\
    \bottomrule
    \end{tabular}%
  \label{tab:quantitative_cmap_comparison}%
\end{table}%

Therefore, we add an additional energy term encouraging local spatial consistency of the weights, defined as
\begin{align*}
    \lossconsist = \frac{1}{\nverts}\sum_{i=1}^{\nverts}\sigma(\neighborhood_{\neighborhoodindex}(\weights_{i})),
\end{align*}
where $\neighborhood_{\neighborhoodindex}(\weights_{i})$ finds the weights of $\neighborhoodindex$ nearest neighbors of vertex $\verts_{i}$. Finally, we find the best weights as
\begin{align*}
    \weights^{*} = \argmin_{\weights}\ \alphadata \lossdata + \alphareg \lossreg + \alphacontrast \losscontrast + \alphaconsist \lossconsist.
\end{align*}
We performed grid search over the loss term weights and number of neighbors $\neighborhoodindex$ on the validation set and eventually set them to $\alphadata=100, \alphareg=0.01, \alphacontrast=100, \alphaconsist=100, \neighborhoodindex=10$.

The final variant is referred to as \cmapconstrastandconsistent{}. As can be seen in \cref{fig:confidence_map_histograms}, the distribution of the weights across the surface is not as extreme as in the case of \cmapcontrast{}, but it is still clearly bi-modal and it discards the lower part of the face as shown in \cref{fig:confidence_map_masks}. This variant yields the best performance, as shown in \cref{tab:quantitative_cmap_comparison} and thus we use it for all experiments in the main paper.

\newpage

\end{document}